# Synthetic Data Generation and Joint Learning for Robust Code-Mixed Translation

**Kartik Kartik**[1], **Sanjana Soni**[2], **Anoop Kunchukuttan**[3],
**Tanmoy Chakraborty**[4], **Md Shad Akhtar**[2]

[1]Washington Post, [2]IIIT Delhi, [4]IIT Delhi, [3]Microsoft India.
kartikaggarwal98@gmail.com, anoop.kunchukuttan@microsoft.com
tanchak@iitd.ac.in, {sanjana19097, shad.akhtar}@iiitd.ac.in

**Abstract**

The widespread online communication in a modern multilingual world has provided opportunities to blend more than one language (*aka* code-mixed language) in a single utterance. This has resulted a formidable challenge for the computational models due to the scarcity of annotated data and presence of noise. A potential solution to mitigate the data scarcity problem in low-resource setup is to leverage existing data in resource-rich language through translation. In this paper, we tackle the problem of code-mixed (Hinglish and Bengalish) to English machine translation. First, we synthetically develop HINMIX, a parallel corpus of Hinglish to English, with ~$4.2M$ sentence pairs. Subsequently, we propose RCMT, a robust perturbation based joint-training model that learns to handle noise in the real-world code-mixed text by parameter sharing across clean and noisy words. Further, we show the adaptability of RCMT in a zero-shot setup for Bengalish to English translation. Our evaluation and comprehensive analyses qualitatively and quantitatively demonstrate the superiority of RCMT over state-of-the-art code-mixed and robust translation methods.

## 1. Introduction

Recent explosion of digital communication around the world has been marked by the growing use of informal language in online conversations. These conversations often feature the *use of words and phrases from multiple languages back and forth into a single utterance*: a phenomenon referred to as code-mixing (CM) or code-switching (Myers-Scotton, 1993a,b; Duran, 1994). *Code-mixing* has become a standard practice both as a form of speech and text in multilingual communities such as Hindi-English (Hinglish), Spanish-English (Spanglish), etc., where people subconsciously alter between languages. Considering it's prominent use, it is imperative to build NLP technologies for code-mixed data.

However, due to the unavailability of annotated data, code-mixing in the domain of text remains largely unexplored. With no official references of CM text in books and articles, online social networks (OSNs) remain the only source of mixed data collection. Further, the real-world unstructured text is highly susceptible to typographical errors and misspellings. These mistakes become more prevalent when languages written in non-romanized scripts such as Japanese, Hindi etc. are adopted to code-mixed scenarios as each word in the originating script can be mapped to multiple probable transliterations, e.g., "*haan bilakul* (bilkul). *yah ek klaasik* (classic) *hai, lekin phir bhee bahut hee ekshan* (action) *aaj ke lie bhee paik* (pack) *hai*" (Yes, definitely. It is a classic, but still very action packed even for today). The problem is exacerbated by the multilingual nature of online code-mixed content, making it essential to understand CM concerning a common language.

---

*The work was carried out when Kartik was a research intern at IIIT Delhi.

Neural Machine Translation (NMT) models have become state-of-the-art in sequence-to-sequence tasks (Sutskever et al., 2014; Bahdanau et al., 2015). At the root of this advancement are two interrelated issues: (i) NMT models need a vast amount of parallel data for satisfactory performance; and (ii) NMT models are brittle to even a slight amount of input noise (Belinkov and Bisk, 2018). In order to circumvent all these challenges, we propose **R**obust **C**ode-**M**ixed **T**ranslation (RCMT) using a joint learning framework. First, to handle the scarcity of code-mixed parallel data, we construct a synthetic Hinglish-English dataset by leveraging a bilingual Hindi-English (Hi-En) corpus. For this, we identify various grammatical patterns in the continuous switching of two languages and formulate a general pipeline for creating a *synthetic CM corpus*.

The generated parallel data is then passed through an *adversarial module* that injects different types of naturally occurring adversarial perturbations to generate a source-side noisy version of the code-mixed dataset. Inspired by multilingual NMT models, we train a joint model for translation of clean and noisy CM text to make the Code-Mixed Translation robust to noisy input. Our experiments show that by *jointly training* both noisy and clean text in a multilingual setting, the model can encode diverse lexical variations of code-mixed words into the shared representation space; thereby, substantially improving the translation quality. Addi-tionally, the need of a parallel CM corpus for every new language pair limits the applicability of NMT models for code-mixed translation. Further, the availability and accuracy of language specific POS-taggers, translation dictionaries, filtering tools become pivotal for building a synthetic CM corpus. To ease this challenge, we propose *zero-shot* CM translation, where a bilingual Bengali-English (Bn-En) parallel corpus is trained along with a code-mixed Hindi-English parallel corpus. This way, the model learns to adapt to the multilingual scenario and translate Bengali CM text to English.

Precisely, the contributions of our work are:
- We formulate a linguistically-informed pipeline for synthetically generating codemix data from parallel non-code-mixed corpora.
- We develop `HINMIX`, the first large-scale **Hin**glish Code-**Mix**ed parallel corpus consisting of ~$4.2M$ parallel sentences. We annotate $2787$ gold standard CM sentences for the evaluation.
- We propose a novel `RCMT` model for effectively translating real-world noisy code-mixed sentences to English.
- We explore *Zero-Shot* Code-Mixed Translation for Bengali code-mixed to English translation without any parallel CM corpus.

**Reproducibility:** Code and datasets are available at https://github.com/LCS2-IIITD/Robust_CodeMIX_MT.

## 2. Related Work

Phenomena of code-mixing and intrasentential code-switching have been fairly studied (Verma, 1976; Joshi, 1982; Singh, 1985). Joshi (1982) proposed a formal framework considering the two language systems and a mechanism to switch between them and further captured essential aspects of intrasentential code-switching. Sankoff (1998) explained the presence of consistent tree labeling, implying a constraint on an equivalence order in constituents around a switch point, whereas, Gardner-Chloros and Edwards (2004) analyzed grammatical rules in code-switching based on various underlying assumptions. Despite a good number of linguistic-grounded code-mixed studies are existing, only a few studies (Dhar et al., 2018; Gupta et al., 2021) have explored it within the translation domain, primarily due to the scarcity of the parallel corpora.

The prevalent usage of CM in day-to-day spoken conversations and online written content has instilled the successful application of CM data in various downstream tasks, such as, POS tagging (Jamatia et al., 2016), sentiment analysis (Patwa et al., 2020), speech recognition (Luo et al., 2018), and machine translation (Dhar et al., 2018). Dhar et al. (2018) initiated the effort to create a 6K pair gold-standard Hindi-English CM dataset. Following this, many researchers proposed various methods for synthetic CM generation – Pratapa et al. (2018) utilized parse trees, whereas, pointer generator network was employed by See et al. (2017) and Winata et al. (2019). Recently, Gupta et al. (Gupta et al., 2020) explored linguistic properties and employed an encoder-decoder model to generate CM sentences automatically without parallel corpus. In another work, Gupta et al. (2021) proposed an mBERT-based (Devlin et al., 2019) technique including alignment to find switch words to convert existing parallel corpus to code-mixed.

The presence of annotated code-mixed data does not ease the target task due to the extensive amount of typos, slang, and phonetic variations in the data; thus, making it implausible to overlook the robustness against the noise of existing solutions. (Belinkov and Bisk, 2018) showed that the model's performance significantly gets affected in the presence of moderate noisy texts. They also provided structure-invariant word representations and robust training on noisy text approaches to boost system performance. In another work, Karpukhin et al. (2019) presented synthetic character-level noise to improve the robustness to natural misspellings for MT systems. However, it lacks to generalize to informal text present on social media discourse. Arguing the need of perturbation-invariant learning, Cheng et al. (2018, 2020) adopted an adversarial stability training objective to learn a perturbation-invariant encoder. Furthermore, Sato et al. (2019) showed promising results by employing adversarial regularization techniques in an NMT model and argued that such methods improve the quality of the translation. An application of adversarial subword regularization (ADVSR) framework was incorporated to expose subword segmentations to regularize NMT models (Park et al., 2020). Although these schemes satisfy the robustness criteria of an NMT model, the nature of noise in CM language largely remains unexplored in Indian languages, which is extremely challenging considering the morphological richness of the language.

Our proposed work is motivated by the gap in research to build an all-inclusive code-mixed translation system that handles the diverse switching nature in CM communities and is robust to a wide range of CM noise. Moreover, the existing works on CM data generation for MT do not guarantee a large-scale dataset. Furthermore, we also explore the zero-shot setting to translate between multiple language-pairs without the necessity to create individual CM datasets.

## 3. Dataset

In this section, we describe the pipeline used to create HINMIX utilizing IITB English-Hindi parallel corpus (Kunchukuttan et al., 2018) – which contains text from TED Talks, Judicial domain, news articles, Wikipedia headlines, etc. Given a source-target sentence pair $S \parallel T$, we generate the synthetic code-mixed data by substituting words in the matrix language sentence with the corresponding words from the embedded language sentence.

**Candidate Word Selection:** We select *nouns*, *adjectives* (JJ), and *quantifiers* to be part of an inclusion list $I$ to identify potential candidates for code-switching. Given a source sentence $S = \{s_1, s_2, \ldots, s_n\} \in L_m$ and a target sentence $T = \{t_1, t_2, \ldots, t_m\} \in L_e$, we obtain POS tags for each word in $S$. Subsequently, we shortlist the candidate words $S = s_i$ such that their corresponding POS tags belong the inclusion list $p_i \in I$. These words can be substituted with their English counterparts $E' = e_j$ to form a code-mixed sentence.

Note that we do not include verbs (VB) and other tags in $I$ as they usually don't follow a one-to-one replacement rule in the code-switched text and often cannot be directly replaced due to the morphological richness of Hindi language. For example, in sentence 'वह खेल रहा है। (He is playing.)', the verb 'playing' is mapped to 'खेल रहा' in Hindi. Choosing either 'खेल रहा' or 'खेल' as potential candidates would result in inaccurate CM sentences ('वह playing है। or वह play रहा है।). For simplicity, we only choose nouns, adjectives, and quantifiers in inclusion list.

**Building Substitution Dictionary:** Once the corpus is POS-tagged using the LTRC parser[1] and candidate words are shortlisted, the substitute words from $L_e$ need to be determined. We propose an alignment-based strategy to build a substitution dictionary. At first, we train an alignment model on IITB Hi-En parallel corpus (Kunchukuttan et al., 2018) to learn word-level correspondence between each parallel sentence. We use the fast-align (Dyer et al., 2013) symmetric alignment model to obtain the source-target alignment matrix. Next, a substitution dictionary $D_i$ for each sentence is obtained, consisting of only words with one-to-one source-target mapping. This approach allows us to deal with the word-sense ambiguity problem by substituting context-dependent foreign words in each sentence, thereby forming a diverse set of code-mixed vocabulary in the corpus.

**Language Switching:** It might appear that the decision to switch a word is a binary choice and that every word in $L_m$ can be replaced from the set of potential substitute words. However, the switching paradigm in a CM utterance depends upon a range of factors such as lexical information available with the speaker, their relative fluency in the languages, speaker's intention to switch, and most importantly, the intrinsic structure of involved languages (Kroll et al., 2008). Hence, instead of substituting every candidate word and generating a single CM sentence, we follow a randomized word-selection and filtering method to obtain multiple CM combinations of a single source sentence.

- **Word Selection:** Given that there can be $2^r - 1$ CM combinations in a sentence of $r$ candidate words, we adopt following length-based heuristics to limit the CM sentences to be generated. This allows us to narrow down the sample space, which otherwise, would have been computationally expensive for large $r$:
  *Heuristic for candidate word selection:*
  – **For r $\leq$ 4**: Use all valid combinations. For example, an $n$-word sentence with 3 candidate words will have $2^3 - 1$=7 CM sentences.
  – **For 5 $\leq$ r $\leq$ 7**: Use $r - 3$ to $r$ candidate word combinations. For example, a sentence with 5 candidate words will have $^5C_2 + ^5C_3 + ^5C_4 + ^5C_5 = 26$ CM sentences.
  – **For r $\geq$ 8**: Use $0.6r$ to $0.7r$ candidate word combinations. For example, a sentence with 15 candidate words will have $^{15}C_9 + ^{15}C_{10}$=8008 CM sentences.
- **Sentence Filtering:** To further narrow down the selection pool and incorporate language structures of bilingual languages into synthetic CM sentences, we use a combination of probabilistic and deterministic NLP evaluation metrics.
  – We use an unsupervised cross-lingual XLM (Conneau and Lample, 2019) model to calculate the perplexity of CM sentences. We observe a good correlation between the fluency of the CM sentence and its perplexity, even when provided with Devanagari Hindi and English text in a single CM sentence.
  – We employ code-mixed specific measures such as Code-Mixing Index (CMI) (Gambäck and Das, 2016) and Switch Point Fraction (SPF) (Gupta et al., 2020) to select sentences between a certain threshold.

Figure 1 shows the process of generating CM sentences in HINMIX. This forms our code-mixed parallel dataset with Hindi (Devanagari)-English CM pairs, $Hi_c$-En. Finally, for each case, we use Google Transliterate API[2] to produce the romanized version of the CM parallel corpora – $Hi_{cr}$-En. In total, we obtain ~4.2M parallel sentences.

---

[1] http://ltrc.iiit.ac.in/analyzer/

[2] https://developers.google.com/transliterate/v1/getting_started

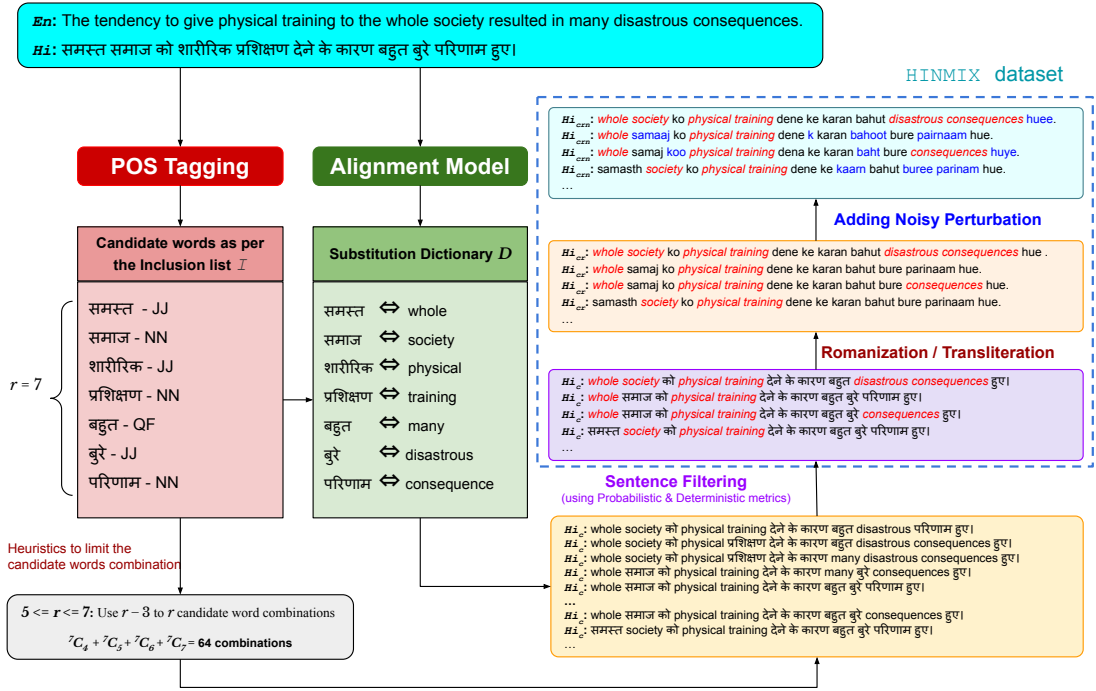

Figure 1: Process of code-mixed sentence generation in HINMIX.

**Human Evaluation:** We report a few samples of HINMIX in Table 1. We can argue that the first two samples are good translation, as both of them are preserving a higher degree of semantics while maintaining the language syntax. We also show the third example which signifies a bad translation primarily due to the wrong POS tag for the word "*khate*" – the word "*khate*" has two common senses, i.e., account (noun) and eating (verb), and the POS tagger misclassify it as a noun instead of a verb. To further assess the quality of our synthetic CM sentences, we perform a human evaluation on 50 randomly selected Hinglish samples. Three bilingual speakers proficient in English and Hindi were asked to rate the adequacy and fluency of each sample on a 5-point Likert scale. The evaluators report the average adequacy and fluency scores of 4.76 and 4.44, respectively.

**Adversarial Module:** The transliteration of non-roman languages depends upon the phonetic transcription of each word, varying heavily with the writer's interpretation of involved languages. With no consistent spelling of a word, it becomes crucial to simulate the real-world variations for the practical application of any CMT model. Hence, we propose to add word-level adversarial perturbations to the transliteration of non-roman words as follows:
- **Switch:** "*t r a n s f e r*" vs "*t r a s n f e r*".
- **Omission:** "*a m a z i n g*" vs "*a m z n g*".
- **Proximity typo:** "*m o b i l e*" vs "*m o v i l e*".
- **Random Shuffle:** "*l a p t o p*" vs "*l o p t a p*".

We inject 30% switch, 12% omission, 12% typo,

| Hi: | पति की प्रेरणा से उन्होंने संस्कृत में लिखित रामायण का बांग्ला में **संक्षिप्त** रूपांतरण किया । (**Pati ki prerana** se unhonne sanskrit men likhit ramayan ka bangla men **sankshipt** rupantar kiya.) |
|---|---|
| En: | At her **husband**'s **persuasion** she translated into Bengali an **abridged** version of the Ramayana from Sanskrit. |
| CM: | Husband ki persuasion se unhonne sanskrit men likhit ramayan ka bangla men abridged rupantar kiya. |
| Hi: | यह **सुरक्षा प्रमाणपत्र विश्वसनीय** नहीं है। (Yeh **suraksha pramanpatra vishvashniye** nahi hai.) |
| En: | This **security certificate** is not **trusted**. |
| CM: | Yeh security certificate trusted nahi hai. |
| Hi: | हम खाने के बाद **आम खाते** थे। (Hum khane ke baad **aam khate** the.) |
| En: | We **ate mangoes** after lunch |
| CM: | Hum khane ke baad mangoes ate the |

Table 1: Samples of generated CM sentences.

and 6% shuffle noise to Hi$_{cr}$ for producing a 60% word-level noisy code-mixed corpus Hi$_{crn}$-En. Both clean (Hi$_{cr}$-En) and noisy (Hi$_{crn}$-En) corpora are further used to train a joint model.

**Development of Gold-standard dataset:** For the gold standard annotation, we take the service of two professional annotators – a male and a female. The annotators are proficient bilingual speakers in the age range of 25-35 years with their first and second languages as Hindi and English, respectively. Given a Devanagari Hindi sentence, annotators were assigned to write the Hinglish conversion that appears as a first thought in the mind. The time-frame for codemix conversion should not exceed 5 seconds once a sentence is read. As there is no standard scheme for roman transliteration of Indic scripts, we ask annotators to transliterate the Devanagari words as per their understand-ing of word structure and its sound pattern. This way the code-mixed sentences are annotated in romanized form with no fixed spelling of any word – as a consequence, same word may take different representations (spelling) in different sentences. This ensures robustness of models as such variation can act as natural noise during testing.

| Statistics | Type | Sentence-level | | | | Token-level | | | | | Char-level | |
|---|---|---|---|---|---|---|---|---|---|---|---|---|
| | | #Sent | #Unique | CMI | SPF | # Hi$_{src}$ | #En$_{src}$ | #EN$_{tgt}$ | Mean | Median | Mean | Median |
| Train | Synthetic | 4.2M | 0.67M | 27.9 | 44.3 | 0.25M | 0.11M | 0.19M | 100.9 | 88 | 18.24 | 16 |
| Dev | Gold | 280 | 280 | 32.6 | 47 | 711 | 667 | 1392 | 65.6 | 64 | 12.17 | 12 |
| Test | Gold | 2507 | 2507 | 32.4 | 45.5 | 4194 | 5923 | 11255 | 124.9 | 111 | 22.8 | 20 |

Table 2: Statistics of HINMIX code-mixed dataset.

**Statistics:** The detailed statistics of the synthetic and gold-standard annotated code-mixed datasets are provided in Table 2. We use the synthetic dataset for the training purpose, while the manually annotated gold dataset is divided into a development set and a test set. In total, there are 4.2M, 280, and 2507 parallel sentence pairs in the train, development, and test sets, respectively.

We also evaluate the complexity of datasets using codemix-specific metrics such as Code-Mixing Index (CMI) and Switch Point Fraction (SPF). CMI measures the percentage of code-mixing in a sentence, whereas SPF computed the percentage of switch-points between the matrix (i.e., Hi) and embedding (i.e., En) language words in a sentence. We observed that both CMI and SPF of synthetic and gold standard datasets have similar scores. It suggests that the synthetically-generated sentences are closely aligned with the manually written CM sentences in terms of the usage of English words and their frequency in a CM sentence.

## 4. Robust Code-Mixed Translation

In this section, we describe our approach for robust translation of CM sentences to English. To capture the context-dependent lexical variations between the noisy and clean corpora, we formulate the cross-lingual translation setting to the code-mixed scenario, referred to as Robust Code-Mixed Translation (RCMT). For this, we jointly train a transformer model in three directions: bidirectional Hindi-English *clean* code-mixed romanized corpus (Hi$_{cr}$⇌En) and Hindi to English *noisy* code-mixed romanized corpus (Hi$_{crn}$→En), where c, r, and n represent the code-mixed, romanized, and noisy versions of a dataset, respectively. We term this setup as RCMT_*roman*.

We employ SentencePiece[3] tokenizer with a unigram subword model (Kudo, 2018) to generate a vocabulary directly from the raw text. As the unigram model calculates subwords according to the occurrence probabilities, directly applying the tokenization to the corpora would result in the under-representation of low-resource languages. Therefore, we undersample the high-resource language by randomly choosing a fixed set of sentences from the corpora to obtain the shared dictionary.

We also define an extended setup, RCMT_*roman+devan*, where we append two non-romanized (or devanagari) code-mixed directions in RCMT_*roman*: bidirectional Hindi-English devanagari corpus (Hi$_c$⇌En). This setup is motivated by the fact that the subwords tokens of Hi$_{cr}$ and Hi$_{crn}$ sentences would contain substantial amount of overlap due to the joint vocabulary. Any noise due to lexical, phonetic, or orthographic variations only perturbs the word at the character level, thereby obtaining similar subwords to some extent. Further, when translating two structurally different sentences (i.e., Hi$_{cr}$ and Hi$_{crn}$ versions of a sentence) to the same target language, the joint model would learn the relationship between those subwords by utilizing their same syntactic and semantic properties. Therefore, the non-canonical nature of noisy text would benefit from the strong implicit supervision of clean sentences even when they are morphologically dissimilar. Since both noisy and clean corpora follow the same origin (Devanagari Hindi), we additionally incorporate two non-romanized code-mixed directions. This modification would enable RCMT to better handle the dependencies among Devanagari and romanized characters besides minimizing the morphological ambiguity across sentences.

**Architecture and Learning Objective:** Inspired by the success of multilingual models, we leverage a sequence-to-sequence joint learning framework to translate code-mixed sentences to English. Unlike typical NMT models trained on a single language pair for one direction, the joint model consists of a single encoder and a decoder for different corpora (code-mixed/romanized/noisy) and directions allowing them to simultaneously learn useful information across language boundaries. For training the joint model from multiple sources to multiple targets (many-to-many), a proxy token for the target language is inserted at the beginning of the source sentence, indicating the intended target at the decoding stage. A high-level architectural diagram of RCMT is illustrated in Figure 2.

The joint model is trained to optimize the sum of categorical cross-entropy (CE) loss with label smoothing (Szegedy et al., 2016) across all language pairs. As our code-mixed datasets are synthetically prepared by replacing words using the matrix language framework (Myers-Scotton, 1993b), learning the model directly using the CE loss would tend to memorize the labels for incorrect source tokens and degrade the model performance. Therefore, we adopt label smoothing to train our proposed model.

---
[3] https://github.com/google/sentencepiece

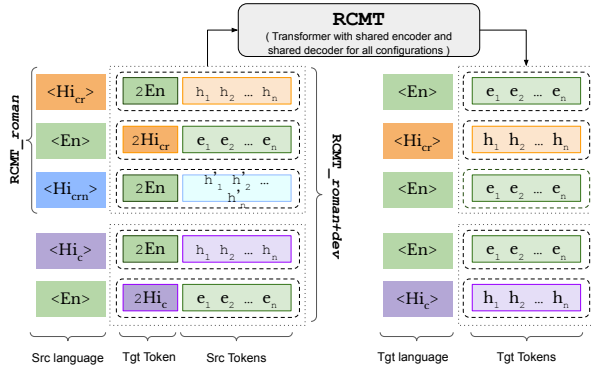

Figure 2: The proposed RCMT model. The subscripts c, r, and n denote codemix, romanized, and noisy version of a dataset. The target token [2T] in the encoder input indicates the intended target language T followed by tokens in the source language S. The target tokens are passed to the decoder sequentially for model training.

## 5. Experiments and Results

We use a standard seq2seq Transformer model (Vaswani et al., 2017) in all our experiments to ensure the same number of parameters. Both encoder and decoder consist of a stack of $6$ identical layers. Each layer comprises a Multi-Head Attention layer with $4$ attention heads and a Feed-forward layer with an inner dimension of $1024$. The shared input and output embedding dimensions are set to $512$. We use a dropout rate of $0.1$, a learning rate of $5 \times 10^{-4}$ and an Adam optimizer with warmup steps of $4000$. A unigram model with character coverage $1.0$ is trained on all languages to obtain a common vocabulary of size $32000$. To implement our model, the fairseq (Ott et al., 2019) toolkit is employed. Finally, we evaluate the quality of models on **SacreBLEU** (Ott et al., 2019) and **METEOR** (Banerjee and Lavie, 2005) metrics.

**Baselines:** We conduct experiments with multiple CM and robust MT baselines for fair comparison of our RCMT approach: • **TFM:** We employ a vanilla Transformer with the same hyperparameters as RCMT for each configuration. • **FCN:** Following (Gehring et al., 2017), we adapt seq2seq fully convolutional network for Robust CMT task. • **mT5:** Xue et al. (2021) put forward a "span-corruption" objective to pre-train a massive multilingual masked LM for sequence generation. • **mBART:** Liu et al. (2020b) used a seq2seq denoising-based autoencoder pre-trained on a large common-crawl corpus. • **PtrGen:** Gupta et al. (2020) used a BiLSTM encoder initialized with XLM feature and pointer generator to decode sentences. • **MTNT:** Vaibhav et al. (2019) proposed to enhance the robustness of MT on the noisy text by pre-training an LSTM model with a clean corpus and fine-tuning it on noisy artificial data. • **MTT:** Zhou et al. (2019) presented a Multi-task Transformer for robust MT that uses dual decoders, one to generate the clean text and another to provide the translation given the noisy input. • **AdvSR:** Park et al. (2020) introduced an adversarial subword regularization scheme for on-the-fly selection of diverse subword segmentation in a sequence resulting in character-level robustness of an NMT model.

To ensure a fair comparison with RCMT, we fine tune each baseline on HINMIX. Moreover, since MTNT, MTT, and AdvSR models are designed for robust (noisy) machine translation, we train them from scratch on HINMIX.

| Model | c | | c+r | | c+r+n | |
|---|---|---|---|---|---|---|
| | B | M | B | M | B | M |
| TFM (Vaswani et al., 2017) | 9.97 | 39.7 | 10.02 | 36.2 | 9.70 | 37.4 |
| FCN (Gehring et al., 2017) | 7.89 | 33.2 | 8.07 | 33.1 | 5.69 | 27.5 |
| mT5 (Xue et al., 2021) | 4.27 | 22.6 | 4.28 | 25.9 | 2.80 | 19.5 |
| mBART (Liu et al., 2020b) | 5.38 | 29.5 | 7.07 | 35.7 | 3.19 | 21.7 |
| PtrGen (Gupta et al., 2020) | 6.51 | 27.18 | 4.68 | 21.15 | 3.04 | 16.1 |
| MTT (Zhou et al., 2019) | - | - | - | - | 10.44 | 38.0 |
| MTNT (Vaibhav et al., 2019) | - | - | 8.48 | 35.1 | 5.92 | 28.0 |
| AdvSR (Park et al., 2020) | - | - | 9.63 | 36.7 | 7.28 | 32.7 |
| **RCMT_roman** | - | - | **13.58** | **45.7** | **11.54** | **41.5** |
| **RCMT_roman+devan** | **13.81** | **46.2** | **13.72** | **45.7** | **11.30** | **40.8** |

Table 3: Comparative results for RCMT on HINMIX. Here, c, r, and n denote codemix, romanized, and noisy version of a dataset. (**B:** SacreBLEU and **M:** METEOR). **Blank entries:** The original MTT, MTNT, and AdvSR models were especially designed for the romanized code-mixed data. Moreover, MTT was specifically designed for noisy codemixed generation.

**Results:** Table 3 presents the results of our robust CMT experiments. We observe that RCMT significantly outperforms all CM and robust MT baselines. Furthermore, we observe minor decline in results with the increase in the corpus/languages (RCMT_${roman}$ → RCMT_${roman+devan}$). We attribute this to the lesser number of parameters for each pair in a joint model when more pairs are added. Regardless, our proposed model handles an all-inclusive CM input (Devanagari, English, romanized, and noisy words) in an efficient manner, thus making it a suitable candidate for practical applications. In the following subsections, we elaborate on the obtained results and their comparisons with the baselines and state-of-the-art systems.

**Code-mixed MT Results:** Seq2Seq models such as transformers (TFM) and convolutional attention networks (FCN) have become the de-facto standard to evaluate MT systems (Liu et al., 2020a; Wu et al., 2019). Following their competitive performance in code-mixed translation tasks (Nagoudi et al., 2021; Appicharla et al., 2021; Dowlagar and Mamidi, 2021), we train individual models in each direction ($Hi_c{\rightarrow}En$, $Hi_{cr}{\rightarrow}En$, $Hi_{crn}{\rightarrow}En$). Table 3 shows the superior performance of TFM over FCN with an avg. improvement of $+2.47$ & $+2.68$ BLEU across CM ($c, c+r$) and robust CM ($c+r+n$) translation models, respectively. A substantial gain of $+3.31B$, $+7.25M$ score (on avg.) over TFM is observed on noisy corpus ($Hi_{crn}{\rightarrow}En$) when it is trained simultaneously with clean corpora ($Hi_{cr}{\rightleftharpoons}En$) in RCMT_roman. Furthermore, the inclusion of Devanagari CM ($Hi_c{\rightleftharpoons}En$) in RCMT_roman+devan improves CM performance; however, it does not provide additional support in the robustness of the system. Also, for $Hi_c \rightarrow En$, RCMT shows stronger results than TFM model even when Devanagari subwords are not shared with any other pair. We hypothesize that training on a common target En enables the encoder to learn overlapping representations for all inputs ($Hi_c, Hi_{cr}, Hi_{crn}$), thereby reducing the effect of script variation and reinforcing the same family correlation.

Previous works in CMT have primarily relied on large-scale multilingual models such as mBART and mT5 (Xue et al., 2021; Liu et al., 2020b; Gautam et al., 2021; Jawahar et al., 2021). For comparison, we adopt the existing approach by finetuning mT5 and mBART models on our CM datasets. Table 3 (row-3 and row-4) highlights the CM performance on these finetuned models. Surprisingly, the romanized code-mixed MT ($c+r$) demonstrates comparable avg. meteor score with $+1.35$% improvement over its Devanagari counterpart ($c$), even though the romanized Hindi text is seen only during finetuning. Conclusively from Table 3, these transfer learning approaches still lag behind RCMT, especially in robust CMT as the pretrained procedure did not involve any kind of CM data. However, it gives us a direction to explore by including CM data in the pre-training steps.

**Robust MT Results:** In order to corroborate the robustness capabilities of RCMT models, we test three noise-robust MT models as baselines: MTT, MTNT, and AdvSR. MTT proves to be most resilient to synthetic noise among other robust baselines with an average BLEU score of $10.44$ against $5.92$ of MTNT and $7.28$ of AdvSR. It could be because it uses a dual decoding scheme to jointly maximize clean text and the translated text. Moreover, the AdvSR model, trained exclusively on noisy corpus, yields better performance than the MTNT model, which is trained on clean corpus $Hi_{cr}{\rightarrow}En$ and finetuned on the noisy corpus $Hi_{crn}{\rightarrow}En$.

In comparison, RCMT reports an average improvement of approx. $+1.0$ BLEU score than MTT (the best baseline model). Furthermore, RCMT_roman has lesser parameter, as compared to MTT, which accounts for increased model size to allocate parameters for the second decoder module. On the other hand, RCMT has the capability to adapt to any number of pairs without increasing the model size. We observed even better performance on meteor scores, where RCMT_roman ($41.5$) and RCMT_roman+devan ($40.8$) report approx $+2.8$ and $+3.5$ better meteor scores than the best baseline, MTT ($38.0$), respectively.

Furthermore, RCMT ($47.9M$) is significantly lighter (in terms of number of parameters) than most of the comparative systems including MTT – AdvSR ($76.9M$), MTT ($120.6M$), FCN ($152.1M$), mT5 ($300M$), and mBART ($680M$). Only MTNT ($21.1M$) and TFM ($43.8M$) have lesser parameters but their performances are not at par with RCMT.

**Generalizability of RCMT:** To further solidify the robustness of our RCMT models, we employ three additional MT datasets: LinCE (Aguilar et al., 2020), SpokenTutorial (Gupta et al., 2021), and IITB Hi-En (Kunchukuttan et al., 2018) datasets. LinCE and SpokenTutorial datasets contain code-mixed sentences, whereas, IITB is a non-CM Hindi-English dataset. Moreover, LinCE contains real-world noisy tweets collected from Twitter, a suitable candidate to assess robustness of the model. We evaluate our trained RCMT models on the test sets of these three dataset. As seen in Table 4, our models obtain better performance across all datasets with avg. BLEU and Meteor scores of $14.17$ and $42.08$, respectively. On LinCE, RCMT models yield comparatively lower scores, possibly due to the higher percentage of noise and the presence of informal tokens (emoticons, hashtags, etc.). Also, our model is able to translate non-CM text with comparable performance as that of CM translations. These results indicate that RCMT performs good on unseen datasets as well.

**Zero-shot Code-mixed MT (ZCMT):** Development of a code-mixed parallel corpus for a new language pair (e.g., Bengalish $\rightleftharpoons$ English) is non-trivial due to various challenges (PoS tagger, alignment model, etc.). Therefore, to negate the limitation of data scarcity, we propose a zero-shot transfer learning approach for code-mix transla-tion in a new language pair. In this approach, we use the previously generated CM corpora to exploit the transfer learning characteristic of cross-lingual models for CMT in an unseen pair. The idea is to utilize the existing non-CM parallel corpus of language $l_1$ and a CM parallel corpus of language $l_2$ for the translation of CM sentences of $l_1$. To this end, we train RCMT with Bengali-English (Bn-En) and Hinglish-English (Hi$_{cr}$-En) parallel corpora. Subsequently, the trained model is employed to convert a code-mixed Bengali (Bn$_c$, Bn$_{cr}$) sentence to English. We argue that the trained model would be able to transfer the code-mixing behaviour onto the network activations in a zero-shot way. We choose Bengali (Bn) due to the availability of both Bn-En large parallel-corpora (Hasan et al., 2020) and Bengali code-mixed SpokenTutorial dataset Bn$_c$-En (Gupta et al., 2021) – it consists of 28K utterances transcribed from code-mixed video lectures. We randomly select 500 and 2000 sentences as the dev and test sets, respectively. The ZCMT is summarized as follows:

- **Training:** Code-mixed Hindi to English [*Devanagari* (Hi$_c$⇌En), *romanized* (Hi$_{cr}$⇌En), *noisy romanized* (Hi$_{crn}$→En)] + Bengali to English [*Eastern-Nagari* (Bn⇌En) and *romanized* (Bn$_r$⇌En)].
- **Testing:** Code-mixed Bengali to English [Bn$_c$→En, Bn$_{cr}$→En]

| Datasets | RCMT_roman | | RCMT_roman+devan | |
|---|---|---|---|---|
| | B | M | B | M |
| IITB (non-CM) | 12.25 | 40.8 | 12.75 | 40.9 |
| SpokenTutorial (CM) | 22.58 | 52.1 | 23.07 | 52.5 |
| LinCE (CM) | 11.06 | 33.9 | 10.28 | 33.5 |
| HINMIX (CM) | 13.58 | 45.7 | 13.72 | 45.7 |

Table 4: Comparison of trained (c + r) RCMT models on other CM and non-CM corpus.

| Model | | Hindi | | Bangla | |
|---|---|---|---|---|---|
| | | B | M | B | M |
| MMT | − | 13.59 | 45.0 | **15.66** | 47.7 |
| | r | 13.05 | 44.1 | 13.83 | 44.3 |
| ZCMT | c | **14.00** | **46.7** | 15.41 | **49.8** |
| | c + r | 13.69 | 46.1 | 14.01 | 47.6 |

Table 5: Comparative performance of RCMT in a zero-shot setting (ZCMT). **Training:** *Bengali-English* (Bn→En) and *code-mixed Hindi-English* (Hi$_{cr}$→En). **Testing:** *Code-mixed Bengali-English* (Bn$_c$→En, Bn$_{cr}$→En).

| | |
|---|---|
| Source (Hi$_{cr}$): | Is **thought** ko sabhi places par support nahin mila. |
| Target (En): | The **concept** is not a universal hit. |
| RCMT_roman | This thought did not support at all the places. |
| Source (Hi$_{cr}$) | Yah aapke relatives aur loved ones ke liye ek **complete** gift hai. |
| Target (En) | It is **perfect** gift for your relatives and loved ones. |
| RCMT_roman | This is a complete gift for your relatives and loved ones |

Table 6: Sample translation of code-mixed (Hi$_{cr}$) sentences to English (En) produced by the proposed RCMT_roman model.

***Results:*** Table 5 shows the effectiveness of zero-shot CM translation ({Bn$_c$, Bn$_{cr}$}→En) by training a joint model using a bilingual Bn-En corpus and our synthetic code-mixed Hi-En corpus. For the baseline model, we test Bn$_c$ and Bn$_c r$ code-mixed translation without training on CM text in a multilingual manner (MMT), i.e., {Hi, Hi$_r$, Bn, Bn$_r$}⇌En + Hi$_{rn}$→En. Interestingly, MMT demonstrates appreciable performance on the Bn test set; however, ZCMT obtains +3.25 improvement on METEOR scores over the MMT model. A possible reason for this can be the nature of the SpokenTutorial (Gupta et al., 2021) test set, which mostly contains technical words and proper nouns as English ($L_e$) words in Bengali ($L_m$) code-mixed text.

Another surprising benefit of our ZCMT model is observed in Hindi CM translation in both Devanagari and romanized texts is that it outperforms RCMT_roman and RCMT_roman+devan scores in Table 3. This indicates that adding languages from the same family (Indo-Aryan) can sometimes improve the code-mixed translation quality despite varying scripts (Devanagari vs. Eastern-Nagari).

**Qualitative Analysis:** Table 6 shows the outputs of a few samples generated through RCMT_roman. We observed that RCMT_roman learns to match the words in source and target sentences – word "*thought*" and "*complete*" are translated as it is from the source sentences to the generated sentences. In the first example, we encounter distorted semantics in the translated text, whereas, a higher degree of semantic content is preserved in the second example, albeit missing a more suitable word "*perfect*". In general, we observed that fluency and adequacy of the generated sentences are encouraging; however, the usage of related or synonym words against the expected words in the generated translation (e.g., "*perfect*" vs "*complete*", "*concept*" vs "*thought*") poses a challenge for RCMT.

## 6. Conclusion

In this work, we proposed a two-phase strategy to translate the real-world code-mixed sentences in multiple languages to English. First, a linguistically informed pipeline was introduced to generate a large-scale HINMIX code-mixed corpora synthetically using a bilingual Hindi-English parallel corpus. Next, we created a perturbed corpus by passing the clean code-mixed corpus to an adversarial module – both of which are simultaneously trained in a joint learning mechanism to learn robust CM representations. Finally, we showed the effectiveness of zero-shot learning on code-mixed MT in Bengali language. Our evaluation showed satisfying performance for both robust Hindi CM and zero-shot Bengali CM translation.

In the future, we would like to extend our work to multiple worldwide code-mixed languages using supervised and unsupervised methods. Additionally, we plan to handle the real-world noise in social media code-mixed texts to further improve the robustness of the system.

## Limitation:

Synthetic data generation always instills a concern regarding the quality of the synthesized sample; however, at the same time, it enables us to generate a wlarge amount of samples in a quick time. Though we did our best to maintain the quality of the dataset in HINMIX, there are few cases of bad translation mainly because of the following reasons:

- **Alignment Errors**: Despite the context-dependent word substitution in HINMIX, it is susceptible to all the alignment errors. Incorrect word mapping between the source-target could completely alter its CM meaning. Also, we substitute words with an only one-to-one correspondence between the source and target, thereby abandoning all words with multiple alignment mapping may have caused issue in appropriate translation in some cases.
- **POS Tagging Errors**: A good POS tagger forms the basis of our code-mixed dataset creation process. In cases a word in the source sentence is incorrectly tagged to a tag in POS inclusion list $I$, its substitute word will not be appropriate. For example in Table 1, the verb "*khate*" gets mistagged to a noun, thereby being replaced by its translation "*ate*".

## Language Resource References

- Kunchukuttan et al. (2018): IIT Bombay Hindi-English parallel corpus.
- Hasan et al. (2020): Bengali-English parallel corpus.
- Gupta et al. (2021): SpokenTutorial parallel corpus.
- Aguilar et al. (2020): LinCE parallel corpus.

## Acknowledgment

Md Shad Akhtar would like to acknowledge the support of SERB-CRG grant and Infosys foundation through Center of AI (CAI)-IIIT Delhi.